\documentclass{article}

\PassOptionsToPackage{numbers, compress}{natbib}
\usepackage[preprint]{neurips_2023}

\usepackage[utf8]{inputenc} % allow utf-8 input
\usepackage[T1]{fontenc}    % use 8-bit T1 fonts
\usepackage{hyperref}       % hyperlinks
\usepackage{url}            % simple URL typesetting
\usepackage{booktabs}       % professional-quality tables
\usepackage{amsfonts}       % blackboard math symbols
\usepackage{nicefrac}       % compact symbols for 1/2, etc.
\usepackage{microtype}      % microtypography
\usepackage[dvipsnames]{xcolor}        % colors
\usepackage{xspace}
\usepackage{subfig, graphicx}
\usepackage{multicol, multirow}
\usepackage{array, wrapfig}
\usepackage[numbers, compress]{natbib}

\newcolumntype{P}[1]{>{\centering\arraybackslash}p{#1}}

\title{Formatting Instructions For NeurIPS 2023}

\author{%
  Bin Xiao$^1$ \quad Chunan Shi$^2$ \quad Xiaonan Nie$^1$$^2$ \quad Fan Yang$^1$ \quad Xiangwei Deng$^2$ \\ 
  \quad
  \textbf{Lei Su$^1$ \quad Weipeng Chen$^1$ \quad Bin Cui$^2$} \\
  $^1$Baichuan Inc. \quad $^2$ Peking University \\
   $\{$xiaobin, yangfan, sulei, chenweipeng$\}$@baichuan-inc.com\\ 
   $\{$spirited\_away, xiaonan.nie, bin.cui$\}$@pku.edu.cn, dengxiangwei@stu.pku.edu.cn\\
}
  
\newcommand{\Sys}{\texttt{Clover}\xspace}

\begin{document}

\title{\Sys: Regressive Lightweight Speculative Decoding with Sequential Knowledge}

\date{}
\maketitle

\begin{abstract}
Large language models (LLMs) suffer from low efficiency as the mismatch between 
the requirement of auto-regressive decoding and the design of most contemporary GPUs.
Specifically, billions to trillions of parameters must be loaded to the GPU cache through its limited memory bandwidth for computation, but only a small batch of tokens is actually computed.  Consequently, the GPU spends most of its time on memory transfer instead of computation.
Recently, parallel decoding, a type of speculative decoding algorithms, is becoming more popular and has demonstrated impressive efficiency improvement in generation. 
It introduces extra decoding heads to large models, enabling them to predict multiple subsequent tokens simultaneously and verify these candidate continuations in a single decoding step. 
However, this approach deviates from the training objective of next token prediction used during pre-training, resulting in a low hit rate for candidate tokens.
% To improve the hit rate and thereby enhance the overall efficiency, 

In this paper, we propose a new speculative decoding algorithm, \Sys, which integrates sequential knowledge into the parallel decoding process. This enhancement improves the hit rate of speculators and thus boosts the overall efficiency.
\Sys transmits the sequential knowledge from pre-speculated tokens via the \texttt{Regressive Connection}, then employs an \texttt{Attention Decoder} to integrate these speculated tokens.  
Additionally, \Sys incorporates an  \texttt{Augmenting Block} that modifies the hidden states to better align with the purpose of speculative generation rather than next token prediction.
We conducted experiments on both Baichuan-Small (with 7B parameters) and Baichuan-Large (with over 100B parameters). 
The results demonstrate that \Sys achieves superior performance compared to existing methods across different model sizes.  
Specifically, \Sys outperforms the baseline by up to $91\%$ on Baichuan-Small and $146\%$ on Baichuan-Large, respectively, and exceeds the performance of the previously top-performing method, Medusa, by up to $37\%$ on Baichuan-Small and $57\%$ on Baichuan-Large, respectively.

\end{abstract}

\section{Introduction}
\label{sec:intro}
Generative large language models (LLMs)~\cite{gpt2,chatgpt,gpt3}, such as GPT, represent a significant breakthrough in artificial intelligence. 
They have demonstrated remarkable proficiency across a diverse range of applications, from composing creative literary works to generating human-like dialogues in chatbots. Their capability to understand and produce language has opened up new avenues for human-computer interaction, automating tasks that necessitate an understanding of context and nuance.

However, despite their strong capabilities, LLMs also present significant challenges related to low generation efficiency on GPUs.
Specifically, these models create text sequentially by generating one output token per step, responding to a user query in two distinct phases: the prefilling phase and the decoding phase.
(1) During the prefilling phase, the model processes all tokens in the input prompt or context in a single iteration to generate the initial output token. 
(2) During the decoding phase, the model, informed by the prompt/context and previously generated tokens, continues to produce subsequent output tokens one at a time through multiple iterations until the response is complete.
Due to the decoding phase involving multiple rounds of generation, where each round processes only a small batch of tokens, the GPUs' computational resources are severely underutilized.
% The majority of the inference time is consumed by the decoding phase.

% The prevailing mechanism used in large language model (LLM) inference, auto-regressive decoding, generates only a single token per request each step. This computation-sparse characteristic severely limits the inference performance of LLM services, as the limiting memory bandwidth on modern GPUs is a significant factor. With the wide deployment of LLM services, the low hardware utilisation in the decoding phase has become a major performance bottleneck that needs to be addressed.

\begin{figure*}[t]
    \subfloat[Medusa Decoding]{
        \centering
        \includegraphics[width=0.37\linewidth]{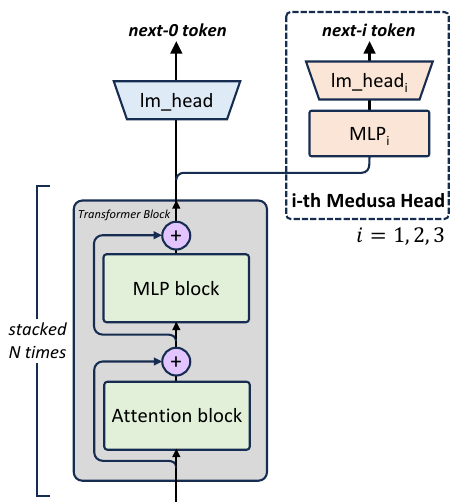}
        \label{fig:medusa}
    }%
    \subfloat[\Sys Decoding]{
        \centering
        \includegraphics[width=0.6\linewidth]{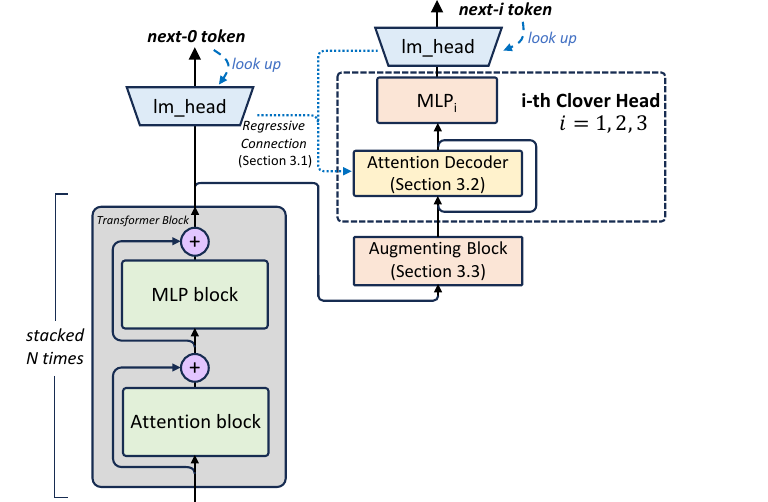}
        \label{fig:seqar}
    }
  \caption{Overview of Medusa decoding and our extended \Sys Decoding.}
\end{figure*}

% Speculative decoding \cite{pmlr-v202-leviathan23a,chen2023accelerating} is one of the acceleration techniques employed to mitigate the performance problem in question. This is achieved by increasing computation density, generating multiple tokens in a single step, while the outputs remain entirely consistent. In particular, the speculative decoding technique requires one or more lightweight draft models to speculate multiple succeeding tokens in negligible overhead. These speculations are then verified by original target model, generating multiple tokens within a single iteration. 
Speculative decoding \cite{pmlr-v202-leviathan23a, chen2023accelerating} is an acceleration technique used to mitigate the performance issues in question. It increases computational density by generating multiple tokens in a single step while ensuring the outputs remain entirely consistent. Specifically, speculative decoding involves one or more lightweight draft models that speculate multiple subsequent tokens with negligible overhead. These speculations are then verified by the original target model, which generates multiple tokens in a single iteration. Speculative decoding generates multiple tokens based on initial speculations. The accuracy of these speculators is critical for decoding speed, while more complex speculators increase inference overhead, subsequently extending latency. Numerous studies \cite{miao2024specinfer, liu2023online, monea2023pass, spector2023accelerating, zhou2024distillspec, zhang2023draft, hooper2024speed} have explored enhancing latency and throughput using independent draft models as speculators. Additionally, recent discussions \cite{cai2024medusa, li2024eagle, bhendawade2024speculative, ankner2024hydra, zhang2024recurrent, zeng2024chimera, du2024glide} have highlighted the advantages of integrated speculators, noting their lightweight nature and ease of deployment.

The Medusa solution \cite{cai2024medusa} leverages lightweight heads as speculators. As shown in Figure \ref{fig:medusa}, it features multiple parallel MLP (Multi-Layer Perceptron) layers that receive inputs from the hidden states of the last transformer block. Each layer is designed to predict a single subsequent token and utilizes a tree-based verification process to simultaneously generate multiple tokens and initiate new speculations. This lightweight head mechanism has led to substantial improvements in inference speed.

% Speculative decoding generates multiple tokens based on the speculation. On the one hand, the accuracy of speculators is a crucial factor in determining the decoding speed. On the other hand, more complex speculators will result in greater inference overhead, which will consequently lead to longer latency. 
% A number of previous studies \cite{miao2024specinfer,liu2023online,monea2023pass,spector2023accelerating,zhou2024distillspec,zhang2023draft,hooper2024speed} have investigated the potential for improving latency and throughput by utilising independent draft models as speculators.
% Integrated speculators have also been discussed in some recent works \cite{cai2024medusa,li2024eagle,bhendawade2024speculative,ankner2024hydra,zhang2024recurrent,zeng2024chimera,du2024glide} due to their lightweight nature and reduced deployment challenges. 

% One of the solutions that employs lightweight heads as speculators is Medusa \cite{cai2024medusa}. As illustrated in Figure \ref{fig:medusa}, it incorporates multiple parallel MLP layers that accept the hidden states from the final transformer block as input. Each layer is trained to predict a single succeeding token, and then applies a tree-based verification process to generate multiple tokens in a single step, as well as new speculations. This lightweight head mechanism has achieved notable inference acceleration.

\begin{figure}{t}
\centering
\includegraphics[width=.4\linewidth]{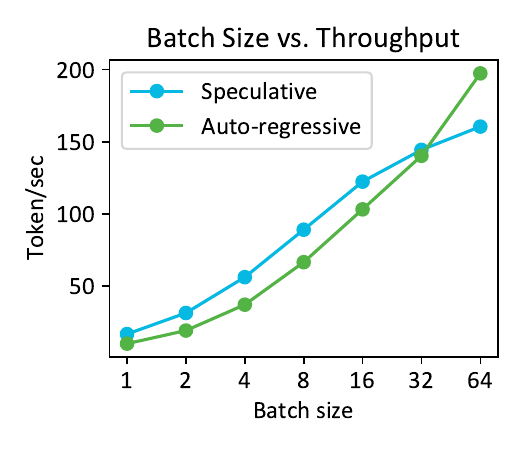}
\caption{Throughput on a model with approximately 30B parameters, supposing speculation length is 5 with 0.4 acceptance rate.}
\label{fig:batch}
\end{figure}

However, Medusa still encounters several challenges that can hinder its performance. Firstly, the Medusa head consists of only a single MLP layer that takes input solely from the final hidden states. Each layer independently speculates on a word at a specified position beyond the next, disregarding the sequential dependencies from previously predicted tokens, which often results in decreased accuracy. Secondly, because the Medusa heads operate independently, the tokens they speculate are combined using a Cartesian product to form an exponentially large token tree. This approach can lead to suboptimal performance when the decoding phase is not constrained by memory, as it generates a surplus of redundant tokens, particularly as the batch size increases. Additionally, the absence of sequential information compromises the effectiveness of the tree pruning algorithm, further impacting performance.

In real-time serving scenarios, where the inference batch size is typically large, speculative decoding often faces computational constraints, leading to performance degradation. Figure \ref{fig:batch} illustrates this trend: speculative decoding substantially outperforms auto-regressive decoding when the number of computed tokens is low. However, as the token count increases, the speedup provided by speculative decoding reaches an inflection point and gradually diminishes due to computational limitations. Consequently, the actual size of the token tree in practice is usually smaller than what is assumed in previous studies.

To address these issues, we introduce \Sys, an enhancement of the Medusa framework. \Sys incorporates a regressive attention block into the speculative phase and introduces the \texttt{Regressive Connection} (Section \ref{sec:rc}), \texttt{Attention Decoder} (Section \ref{sec:ad}), and \texttt{Augmenting Block} (Section \ref{sec:ab}). These components enable speculators to utilize additional sequential knowledge, enhancing their accuracy. Moreover, the regressive architecture not only improves the precision of the speculations but also generates a token tree with more comprehensive dependency information.

We evaluate \Sys in a setting that more closely resembles the real scenario, which involve various larger batch sizes and a smaller token tree. The results on Baichuan model family show that \Sys method achieves a maximum throughput improvement of $2.56\times$ throughput improvement over vanilla decoding and $1.25\times$ - $1.43\times$ over Medusa decoding. 
Moreover, \Sys demonstrates an $11.7\%$ - $26.4\%$ improvement in accuracy on speculative heads, with a particularly notable increase of over $20\%$ in the latter heads. Additionally, it generates $50\%$ - $76\%$ more extra tokens (except the first) per step than the Medusa method, thanks to the regressive mechanism.

To summarize, our contributions can be outlined as follows:

\begin{itemize}
\item We propose \Sys, a new speculative decoding algorithm which incorporates an additional auto-regressive attention block to facilitate the consideration of sequential knowledge.
% an extension based on Medusa head, incorporating an additional auto-regressive attention block to facilitate the consideration of sequential knowledge.
% \item We investigate potential straightforward modifications to the model architecture with the objective of enhancing sequential knowledge from the entire inputs. Furthermore, we introduce an additional block into \Sys with the intention of improving performance. 
\item We introduce three key components to improve the original parallel decoding algorithms, including the \texttt{Regressive Connection} for utilizing sequential information from previously speculated tokens, the \texttt{Attention Decoder} for combining the speculated tokens with current inputs, and the \texttt{Augmenting Block} for modifying the hidden states to better align with the purpose of speculative generation.
\item Evaluations are conducted on both Baichuan-Small (with 7B parameters) and Baichuan-Large (with over 100B parameters). And results show that our \Sys achieved better efficiency compared to existing methods, such as Medusa.
% \item We employ \Sys in a manner analogous to the actual scenario to conduct evaluation as well as other decoding algorithms. The results still reveal that \Sys exhibits considerable accuracy improvement on all heads over Medusa method, as well as throughput improvement over both Medusa decoding and auto-regressive decoding.
% \item We design a dynamic programming algorithm based on regressive speculations to construct optimal token tree within given maximal tree size constraint, significantly decrease the tree size without unexpected accuracy loss.
\end{itemize}

\section{Background}
% \CA{===Thinking to remove this section, duplicated to the intro section.===}
\subsection{Speculative Decoding}
\begin{figure*}[th]
    \subfloat[Auto-regressive Decoding]{
        \centering
        \includegraphics[width=0.42\linewidth]{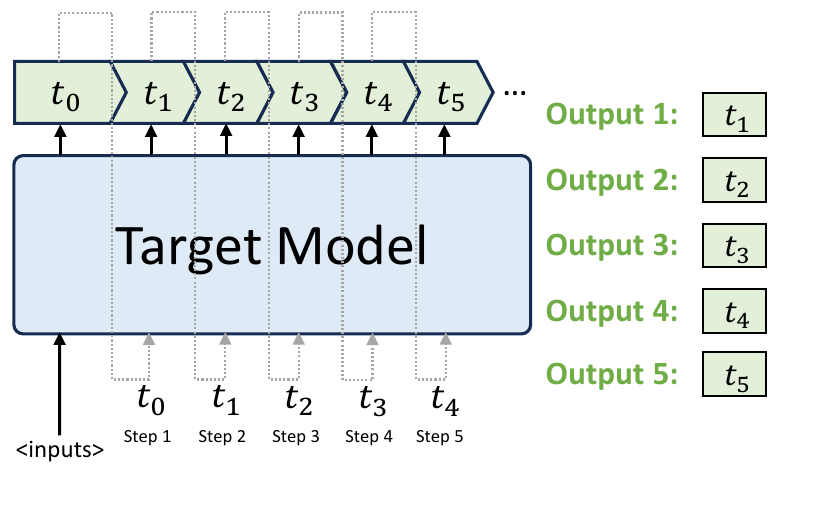}
        \label{fig:ar_dec}
    }%
    \subfloat[Speculative Decoding (maximal speculation length is 4)]{
        \centering
        \includegraphics[width=0.55\linewidth]{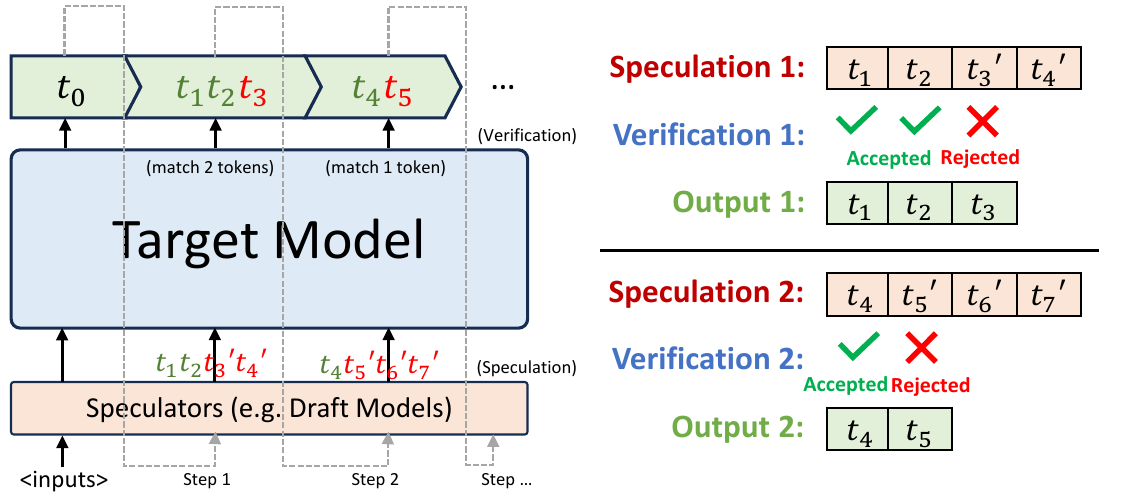}
        \label{fig:sp_dec}
    }
  \caption{The comparison between Auto-regressive Decoding and Speculative Decoding. Speculative Decoding may generate multiple tokens in a single step based on the speculation, thus achieves less decoding iteration and lower inference latency.}
\end{figure*}
\begin{figure}[th]
\centering
\includegraphics[width=0.98\linewidth]{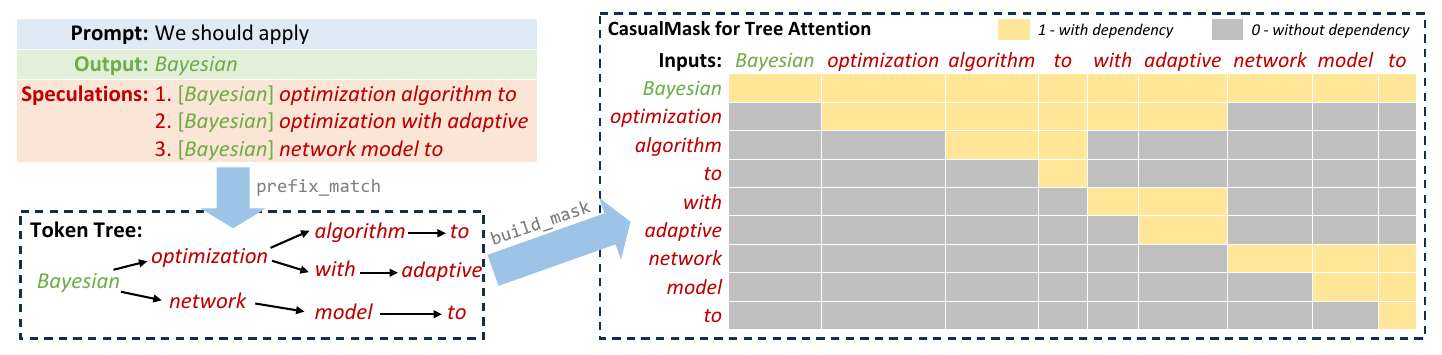}
\caption{A demonstration of Tree Attention in Speculative Decoding. Multiple speculations are merged by prefix matching to form a tree, and its topology dependency is represented in a 2-D matrix as the casual mask in Attention computation.}
\label{fig:tr_aten}
\end{figure}

Speculative decoding \cite{pmlr-v202-leviathan23a,chen2023accelerating}, depicted in Figure \ref{fig:sp_dec}, is an advanced technique that accelerates LLM inference by leveraging hardware computational resources more efficiently. This method distinguishes itself from traditional auto-regressive decoding by calculating and generating multiple tokens simultaneously in each iteration.

At the core of speculative decoding lies a speculator component, usually a smaller model often referred to as the draft model, which predicts several subsequent tokens. This approach contrasts with auto-regressive decoding, where only the last generated token is fed into the system. In speculative decoding, the original LLM (the target model) receives all speculated tokens as input. This allows the target model to compute attention scores and derive logits over multiple tokens effectively, ensuring that it can generate consistent outputs within a single iteration. This stage is termed the \textit{verification phase}, during which the target model screens out any incorrect tokens from the speculations. As a result, speculative inference can produce equivalent outputs with fewer decoding steps, thereby enhancing latency efficiency.

% Speculative inference \cite{pmlr-v202-leviathan23a,chen2023accelerating} represents an acceleration technique for LLM inference, as depicted in Figure \ref{fig:sp_dec}. In comparison to auto-regressive decoding, speculative inference is characterised by a more efficient utilisation of the computational resources of the hardware, as it involves the computation and generation of multiple tokens in each iteration. 

% Speculative inference relies on an additional speculator component to generate predictions about multiple succeeding tokens, which is usually represented by a small model (also known as draft model) in most speculative decoding system. In contrast to auto-regressive decoding that only fed with the last generated token, the inputs of original LLM (i.e. target model) in speculative decoding are the entire speculation. The target model is able to compute attention scores and fetch logits over multiple tokens thanks to the speculation. It is also be able to generate outputs with consistency guarantees in only one iteration. 
% This forward process is also known as the \textit{verification phase}, as the target model rejects wrong tokens in the speculation. Speculative inference generates the same outputs with fewer decoding steps, thus achieving latency improvement.

\subsection{Tree Attention}

Tree Attention \cite{miao2024specinfer} is utilized to calculate attention scores for multiple speculations in parallel. By applying prefix matching to various speculated sequences, the speculation results are organized into a \textit{token tree}, which is represented as a 2-D matrix (Figure \ref{fig:tr_aten}).

It is important to note that the attention block is the only component within the modern LLM architecture that requires knowledge of sequential dependency. The scoring of tree-structured tokens is a relatively straightforward task and can be achieved by configuring the attention's Causal-Mask to align with the topological matrix. Tree Attention facilitates the integration of multiple speculations with minimal computational overhead, a feature widely implemented in many speculative decoding systems such as \cite{he2024rest, yun2021spectr, xu2023llmcad}.

% Tree Attention \cite{miao2024specinfer} is employed to compute the attention scores of multiple speculations in parallel. After applying prefix matching to multiple speculated sequences, the entire speculation results can be constructed as a \textit{token tree}, which can be described as a 2-D matrix (Figure \ref{fig:tr_aten}).

% Note that the attention block is the only component that requires sequential dependency knowledge in the modern LLM architecture. The computation of the score for tree-structured tokens is a relatively straightforward process, which can be implemented by setting the attention's Causal-Mask to the topological matrix. Tree Attention allows the integration of multiple speculations while imposing a minimal computational burden, a feature that has been widely adopted in the majority of speculative decoding systems, such as \cite{he2024rest,yun2021spectr,xu2023llmcad}.

\subsection{Medusa Decoding}
% Figure \ref{fig:medusa} depicts Medusa \cite{cai2024medusa}, which introduces certain independent and parallel MLP heads. The $i$-th head is fine-tuned to predict the next-$i$ token behind the actual output token in each iteration. Such lightweight heads consist of Medusa's speculator, integrated in the target model to perform speculation and verification simultaneously. 
Figure \ref{fig:medusa} illustrates the Medusa architecture \cite{cai2024medusa}, which features several independent and parallel MLP heads. Each of these heads, designated as the $i$-th head, is specifically fine-tuned to predict the next-$i$ token following the actual output token during each iteration. These lightweight heads constitute the speculator component of the Medusa system, seamlessly integrated into the target model. This integration allows for simultaneous speculation and verification within the decoding process. The design of these heads enables Medusa to effectively manage the balance between computational efficiency and predictive accuracy, ensuring that each token generated contributes optimally to the overall sequence coherence and context relevance.

% \Sys is an extension based on such lightweight integrated heads, introducing regressive connections to make more comprehensive use of sequence dependency knowledge.
\section{Clover Design}
\begin{figure}[t]
\centering
% \vspace{-0.5cm}
\includegraphics[width=.78\linewidth]{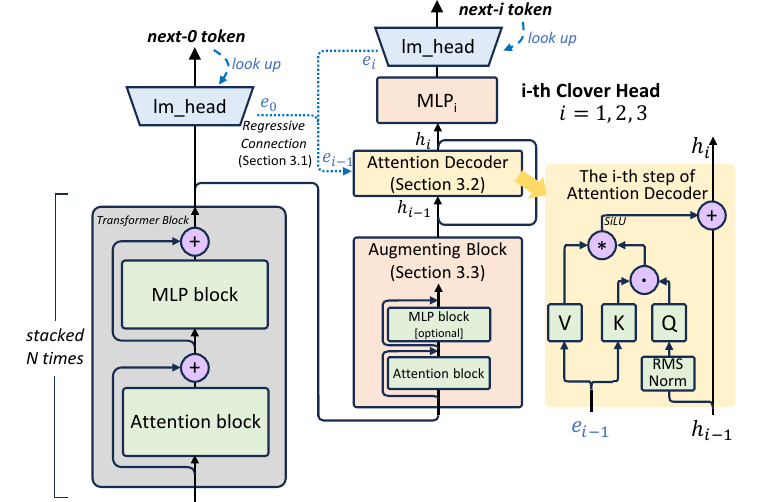}
% \vspace{-0.1cm}
\caption{Detailed architecture design of \Sys.}
\label{fig:seqar_full}
% \vspace{-0.3cm}
\end{figure}

Figure \ref{fig:seqar_full} shows how \Sys is integrated into existent LLM as the speculator. \Sys introduces three incremental components to leverage sequential knowledge: 
\texttt{Regressive Connection}, \texttt{Attention Decoder} and \texttt{Augmenting Block}. 
The Regressive Connection enables sequential dependency from preceding speculated tokens to be considered when a speculator generating the next token. The Attention Decoder is the factual regressive block in \Sys, combining the hidden states from the last transformer block and previously speculated token, merging sequential knowledge between pre-speculated tokens and the entire input sentence. While the Augmenting Block is an additional transformer or self-attention block appended to the target model, used for enhancing sequence features to improve speculator accuracy.

\subsection{Regressive Connection}
\label{sec:rc}
Each Medusa head is responsible for speculating the token at the specified location, without considering the pre-generated speculation, as shown in Figure \ref{fig:medusa}. Although such independence enables multiple heads compute in parallel , the neglect of sequence dependencies limits the hit rate of speculation, and further increases the inference latency. 

\Sys applies regressive connection to the speculator, depicted as the blue dotted lines in Figure \ref{fig:seqar_full}. The embedding vectors of current speculated tokens will be regressively used to predict the token at the next position. Introducing such sequential dependency knowledge offers two benefits: Firstly, speculative heads are able to generate predictions more accurately with previous tokens known, thus decrease inference latency. Although the critical path of the computation becomes proportional to the depth of the speculation and loses a certain amount of parallelism, the increase in speculation accuracy rather improves the overall latency. 

Secondly, since every speculated token in the latter position has one token in its previous location as the precursor, the token tree for verification phase can have greater information density. In contrast to the exponentially sized token tree that result from the independence of words at each position, smaller token tree with sequence-dependency information is easy for pruning and less likely to meet computation bound on modern GPUs, while introducing negligible information loss.

\subsection{Attention Decoder}
\label{sec:ad}
\Sys introduces cross attention decoder as the actual regressive block. The decoder takes two vectors as inputs: the embedding vector from the previous token, and the hidden states throughout the speculation. Specifically, considering the computation flow on the $i$-th head, to generate the next-$i$ token (denoted as $\mbox{tok}_i$), the inputs of cross attention decoder are: the normalized embedding vector of the token $\mbox{tok}_{i-1}$ (denoted as $e_{i-1}$), and the hidden states from the last speculation step (denoted a $h_{i-1}$). The computation can be formulated as follows:
\begin{equation}
Q_{i}=W_Q\cdot normalize(h_{i-1}),~K_{i}=W_K\cdot e_{i-1},~V_{i}=W_V\cdot e_{i-1},
\end{equation}
\begin{equation}
h_{i}=h_{i-1} + Attention(Q_i,K_i,V_i),
\end{equation}
, where $h_i$ is the output of the cross attention decoder, fed into the corresponding MLP layer to generate $\mbox{tok}_{i}$. For the first head, the hidden states $h_0$ comes from the last transformer block of the target LLM model (or the Augmenting Block, see below), and the embedding $e_0$ is from the next-0 token $t_0$ generated by the target model. 

The hidden states $h_i$ is recursively propagated throughout the entire speculation phase, piggybacking the features from the entire input sentence. The role of \Sys's Attention Decoder is combining and resolving the information from both input sentence and the previous speculated tokens, assisting the succeeding MLP layer to speculate token at present position with more sequential knowledge.
We also explore the effectiveness of using the MLP layer as a regressive block, but get sub-optimal performance (more details in Section \ref{sec:ablation}), this is probably because simply concatenating the two input vectors makes it harder to learn and extract valid features. \Sys's Attention Decoder has negligible overhead due to the fact that the inputs are only two vectors per request or beam.

\subsection{Augmenting Block} 
\label{sec:ab}
The original target LLM is pre-trained for just predicting the next token. To extract more information for speculators to predict more succeeding tokens, we append an additional transformer block to augment features from the entire input sentence. The output of this additional augmenting block (i.e. $h_0$) is fed into Attention Decoder. Introducing such a whole layer incurs just a small computation overhead (e.g. approximately $1/N_{layer}$ of inference time), while the accuracy gain from the augmenting block outweighs the time it consumes.

We explore different architectures to build this augmenting block,and find the phenomenon that the attention block contributes the largest accuracy gain to all the speculative heads. We also notice that the MLP block only adds approximately 1\% accuracy gain, so we leave the MLP layer in Augmenting Block optional. We still add the MLP block in our \Sys implementation since it does indeed increase accuracy, while incurring only negligible overhead.
% Moreover, we also attempted to append this augmenting block to the middle transformer block, but found limited improvement in this method. 
Concerning evaluation results are discussed in Section \ref{sec:ablation}.

\subsection{Other Details}

Each medusa head equipped with an individual LM head, containing a large amount of parameters (i.e. the hidden size multiplies the vocabulary size) and make it more time-consuming for training. In \Sys, all the speculative heads share the original LM head in the target model. Furthermore, in regressive connection, the embedding vector of last generated token is given by LM head as well (the look up arrow in Figure \ref{fig:seqar}). 
Specifically, the embedding vector $e_i$ is given by: the one-hot vector of token $t_i$ multiplied by the transposed normalized weight matrix in the LM head.
Compared with looking up from the embedding table, we believe such embedding distribution is much closer to the hidden states from  the last transformer block, where the weights are used to initialize the augmenting block, reducing the difficulty of fine-tuning.

\section{Evaluation}
% How to finetune: hyper-params, dataset, initial weights, epoch, ...

\subsection{Experiment Settings}
% Hardware, engine, models, configuration (batch size, temperature, ...), metrics (token/step, token/sec), baselines, ...

\paragraph{Models and baselines} Both the Medusa and \Sys approaches are employed on the \texttt{Baichuan Small} (with 7B parameters) and \texttt{Baichuan Large} (with over 100B parameters) models \cite{yang2023baichuan} with the number of lm head is 3, named as Medusa(Baichuan) and \Sys{Baichuan}, respectively. 
In order to ensure the fairness of the comparison, the same inference engine, tree construction and tree sampling algorithm
% \footnote{Currently, we apply the tree searching algorithm for both Medusa and \Sys architecture presented in Medusa technical report \cite{}, since construction algorithm in section \ref{sec:construct_dp} can not be directly applied to Medusa.} 
are used for all scenarios. We also evaluate auto-regressive decoding under the same circumstances. 

\paragraph{Dataset} We employ the Baichuan internal supervised fine-tuning (SFT) dataset, containing approximately 0.15B tokens, $95\%$ of which are Chinese, to train both Medusa(Baichuan) and \Sys(Baichuan). We then evaluate inference performance on another internal Baichuan dataset, which consists of a variety of tasks: retrieval augmentation(\textbf{RA}), multi-turn conversation(\textbf{MC}), code(\textbf{Code}), information process(\textbf{IP}), creation(\textbf{CA}), logical reasoning(\textbf{RS}), math(\textbf{Math}), tabular(\textbf{Tab}), question answering(\textbf{QA}) and medical suggestion(\textbf{Med}). Each of the ten tasks contains 100 dialogues.

\paragraph{Training} Both models are trained with all weights frozen in the target model . For Medusa(Baichuan), the initial weight settings correspond to the configuration given in the Medusa technical report \cite{cai2024medusa}. While for \Sys(Baichuan), the initial weights in the Augmenting Block are identical to the last transformer block in the target model, and the initialization of the MLP layer is the same as in Medusa's method. For the Attention Decoder, the weights of Q and K are initialized with identical matrix with Gaussian noise added, while the V matrix is set to all zero. We train the heads for 1 epoch, with  $(\beta_1=0.9,\beta_2=0.999)$ for the AdamW optimizer. The learning rate\footnote{Cosine decay is applied to the learning rate.} is set to 1e-3 for \texttt{Baichuan Small}, and 6e-4 for \texttt{Baichuan Large}. For both models equipped with \Sys, the trainable parameters are approximately 0.2B and 2B, taking 2 hours to train on 8x A800 NVIDIA GPU and 32x H800 NVIDIA, respectively.

\paragraph{Metrics} We choose \texttt{tokens/step} and \texttt{tokens/second} as our main metrics, followed by prior speculative decoding works. The former metric measures the accepted length, indicating the accuracy of speculators, while the latter metric reports the overall system throughput. We report top-k accuracy of each head in ablation study to gain more intuitive insight into diverse model architecture.

\subsection{End-to-end Results}
\begin{figure}[th]
    \centering
    \includegraphics[width=0.8\linewidth]{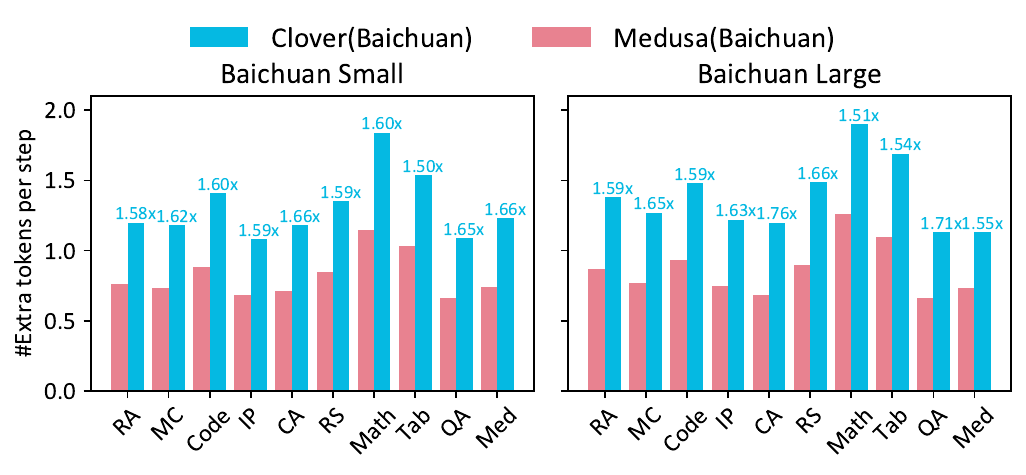}
    \caption{Number of extra generated tokens (excluding the first one) per step on various tasks.}
    \label{fig:tokperstep}
\end{figure}

\begin{table}[t]
    \centering
    \small
    \begin{tabular}{|P{1cm}|P{0.9cm}|P{1.3cm}|P{1.0cm}P{1.0cm}P{1.0cm}P{1.0cm}P{1.0cm}P{1.0cm}|}
        \hline
        Model & \multirow{2}{*}{Task} & \multirow{2}{*}{Approach}  & \multicolumn{6}{c|}{Tokens/second and Improvement over Vanilla Decoding} \\
        Size&&&bs=4 & bs=8 & bs=16 & bs=24 & bs=32 & bs=48\\\hline\hline 
% ============== Auto-gen Begin ==============
% processing  CA  for  Small  model...
% clover/medusa: [1.22161242 1.0858777  1.11055674 1.11981138 1.09177921 1.08537868]

\rule{0pt}{2.6ex}
        \multirow{12}{*}{Small} & \multirow{6}{=}{\centering\textbf{CA}} & \multirow{2}{=}{\centering \Sys(Baichuan)}  &\textbf{195.9}&\textbf{373.1}&\textbf{615.5}&\textbf{872.0}&\textbf{1035.2}&1352.3\\
        &&&\textbf{+44\%}&\textbf{+40\%}&\textbf{+22\%}&\textbf{+19\%}&\textbf{+14\%}&-39\%\\[1ex]%\cline{3-10}
        && \multirow{2}{=}{\centering Medusa (Baichuan)}&160.3&343.6&554.3&778.7&948.2&1246.0\\
        &&&$+18\%$&$+28\%$&$+10\%$&$+6\%$&$+4\%$&$-43\%$\\%\cline{3-10}
        && \multirow{2}{=}{\centering Vanilla}&\multirow{2}{*}{135.5}&\multirow{2}{*}{266.4}&\multirow{2}{*}{501.3}&\multirow{2}{*}{731.6}&\multirow{2}{*}{903.2}&\multirow{2}{*}{\textbf{2217.4}}\\ 
        &&& &&&&&\\\cline{2-9}

% processing  Math  for  Small  model...
% clover/medusa: [1.23739663 1.20136255 1.04357991 1.25633109 1.16758405 1.09715834]

\rule{0pt}{2.6ex}
         & \multirow{6}{=}{\centering\textbf{Math}} & \multirow{2}{=}{\centering \Sys(Baichuan)}  &\textbf{232.2}&\textbf{411.3}&\textbf{673.3}&\textbf{988.4}&1137.7&1462.8\\
        &&&\textbf{+91\%}&\textbf{+76\%}&\textbf{+57\%}&\textbf{+67\%}&-5\%&-21\%\\[1ex]%\cline{3-10}
        && \multirow{2}{=}{\centering Medusa (Baichuan)}&187.6&342.4&645.2&786.8&974.4&1333.2\\
        &&&$+54\%$&$+46\%$&$+50\%$&$+32\%$&$-18\%$&$-28\%$\\%\cline{3-10}
        && \multirow{2}{=}{\centering Vanilla}&\multirow{2}{*}{121.5}&\multirow{2}{*}{233.5}&\multirow{2}{*}{428.5}&\multirow{2}{*}{591.6}&\multirow{2}{*}{\textbf{1202.4}}&\multirow{2}{*}{\textbf{1874.0}}\\ 
        &&& &&&&&\\\hline

% processing  CA  for  Large  model...
% clover/medusa: [1.27920141 1.30641846 1.30834777 1.31149193 1.29792143 1.47060784]

\rule{0pt}{2.6ex}
        \multirow{12}{*}{Large} & \multirow{6}{=}{\centering\textbf{CA}} & \multirow{2}{=}{\centering \Sys(Baichuan)}  &\textbf{169.3}&\textbf{290.2}&\textbf{488.5}&\textbf{638.4}&\textbf{754.4}&\textbf{938.2}\\
        &&&\textbf{+86\%}&\textbf{+55\%}&\textbf{+34\%}&\textbf{+23\%}&\textbf{+22\%}&\textbf{+5\%}\\[1ex]%\cline{3-10}
        && \multirow{2}{=}{\centering Medusa (Baichuan)}&132.3&222.1&373.3&486.8&581.3&638.0\\
        &&&$+45\%$&$+19\%$&$+2\%$&$-5\%$&$-5\%$&$-28\%$\\%\cline{3-10}
        && \multirow{2}{=}{\centering Vanilla}&\multirow{2}{*}{90.7}&\multirow{2}{*}{186.2}&\multirow{2}{*}{362.8}&\multirow{2}{*}{515.5}&\multirow{2}{*}{615.8}&\multirow{2}{*}{887.8}\\
        &&& &&&&&\\\cline{2-9}

% processing  Math  for  Large  model...
% clover/medusa: [1.3029528  1.2678733  1.36922368 1.28433919 1.23886967 1.36579433]

\rule{0pt}{2.6ex}
         & \multirow{6}{=}{\centering\textbf{Math}} & \multirow{2}{=}{\centering \Sys(Baichuan)}  &\textbf{207.3}&\textbf{342.1}&\textbf{549.5}&\textbf{715.4}&\textbf{874.3}&\textbf{1067.8}\\
        &&&\textbf{+146\%}&\textbf{+103\%}&\textbf{+81\%}&\textbf{+62\%}&\textbf{+63\%}&\textbf{+36\%}\\[1ex]%\cline{3-10}
        && \multirow{2}{=}{\centering Medusa (Baichuan)}&159.1&269.8&401.3&557.0&705.7&781.8\\
        &&&$+89\%$&$+60\%$&$+32\%$&$+26\%$&$+31\%$&$+0\%$\\%\cline{3-10}
        && \multirow{2}{=}{\centering Vanilla}&\multirow{2}{*}{84.2}&\multirow{2}{*}{168.3}&\multirow{2}{*}{302.9}&\multirow{2}{*}{440.9}&\multirow{2}{*}{535.2}&\multirow{2}{*}{780.9}\\
        &&& &&&&&\\\hline

% ============== Auto-gen End ==============
    \end{tabular}
    \vspace{10pt}
    \caption{End-to-end throughput on \texttt{Baichuan Small} and \texttt{Baichuan Large} with different decoding methods on two tasks, where \texttt{bs} in the head means batch size, and Vanilla refers to auto-regressive decoding. Results for other tasks are shown in Appendex \ref{sec:ap_e2e}.}
    %  \CA{To Nolan, please give some suggestion to make this table attractive.}
    \vspace{-10pt}
    \label{tab:e2e}
\end{table}

We evaluate the end-to-end performance at different batch sizes. As mentioned in Section \ref{sec:intro}, in real-time serving environment, system need to compute requests with large batch sizes and easily meet the computational bound. Thus we set token tree size to 4 for both speculative decoding methods. We also investigate and find that further expansion of the tree sampling size leads to marginal effects (see Appendix \ref{sec:tree_size}).

Figure \ref{fig:tokperstep} illustrates the average number of tokens generated per step for \Sys and Medusa methods on different tasks. Note that the value on the vertical axis is the \textbf{extra} tokens per step, excluding the actual token generated by target model, which more accurately reflects the performance of the speculator. \Sys generates 50\% - 76\% more extra tokens per step than Medusa method on all tasks, highlighting its superiority over Medusa architecture in terms of speculator accuracy.

The end-to-end throughput (i.e. \texttt{tokens/second}) results are shown in Table \ref{tab:e2e}. Both \Sys and Medusa speculative decoding methods outperform auto-regressive decoding (at most $2.05\times$ - $2.56\times$ in terms of throughput on \texttt{Baichuan Large} model) due to effective hardware utilisation in most scenarios. We also find that the advantage of both speculation decoding methods generally diminishes with increasing batch size. This is because speculative decoding with larger batch size is getting closer to the computational bounded. The performance fluctuation is due to unpredictable random factors during the inference and a not fully optimized implementation of our engine.
More results for other tasks can be found in Appendix \ref{sec:ap_e2e}, \Sys(Baichuan) still retains its best performance over Medusa(Baichuan) and auto-regressive decoding in all categories.

Moreover, \Sys decoding generates more tokens per step and achieves higher throughput than Medusa decoding in all scenarios (at most $1.26\times$ and $1.47\times$ for \texttt{Baichuan Small} and \texttt{Baichuan Large}, respectively\footnote{The value is the ratio of \Sys to Medusa throughput.}), because the gain in head accuracy from the additional components proposed in Clover outweighs their computational overhead. 
The advantages of our system over Medusa are even more pronounced for larger model sizes, as the speculator module makes up a smaller proportion of the overall model.
The sequential knowledge from pre-generated speculation tokens helps the current head to predict next speculation token more accurately, especially when speculating a long multi-token phrase that appears first in the first head, but not at the next-0 token (i.e. the actual output token).

% We also evaluate acceptance length for both \Sys and Medusa decoding on various types of tasks, the results are depicted in \CA{Appendix.}

\subsection{Ablation Study}
\label{sec:ablation}
\begin{figure*}[ht]
    \subfloat[Ablation Study on Components]{
        \centering
        \includegraphics[width=0.49\linewidth]{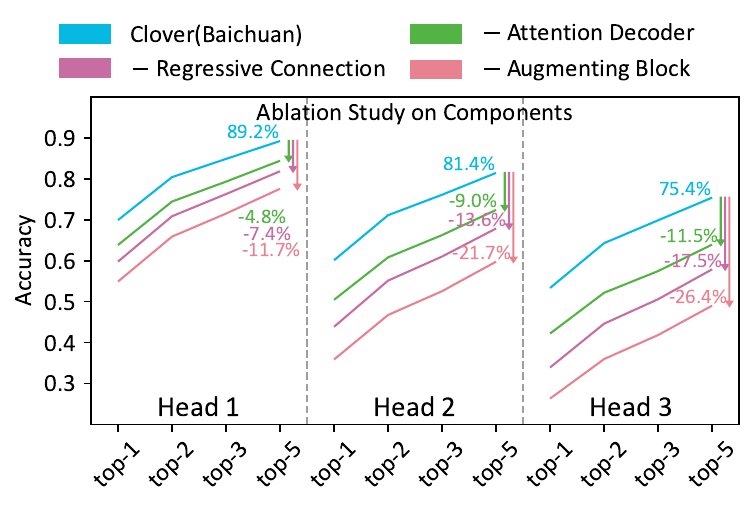}
        \label{fig:ab_cp}
    }%
    \subfloat[Exploration on Augmenting Block]{
        \centering
        \includegraphics[width=0.49\linewidth]{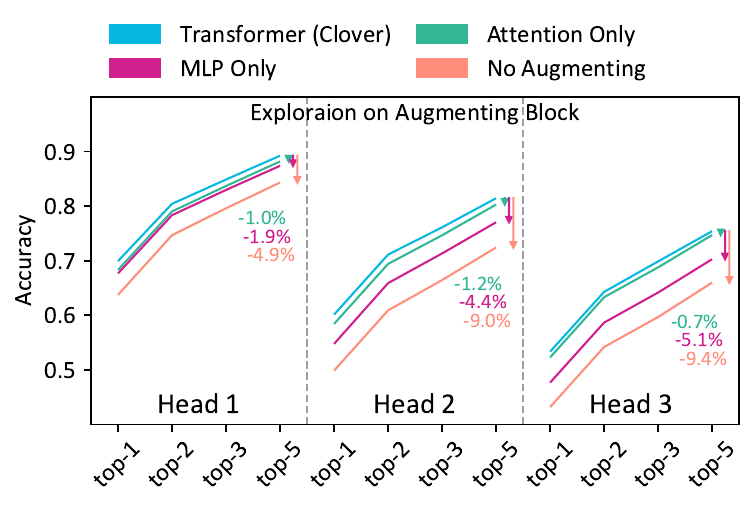}
        \label{fig:ab_ag}
    }
  \caption{Ablation study on \texttt{Baichuan Small} model.}
\end{figure*}

In Ablation study, we use top-k accuracy of each head as the metrics to intuitively understand how each component affects accuracy.
\paragraph{Ablations on Components}
We start from complete \Sys(Baichuan) and gradually remove: Attention Decoder, Regressive Connection and Augmenting Block. Note that after removing all the three components, the model architecture becomes identical to the Medusa(Baichuan). Figure \ref{fig:ab_cp} shows the accuracy of \Sys(Baichuan) with difference components enabled.

By removing Attention Decoder, taking the MLP layer as the regressive block instead, the top-5 accuracy of the three heads reduces by $(4.8\%,9.0\%,11.5\%)$. This is because MLP layer itself will not distinguish different embedding vectors, make it hard to learn valid sequential knowledge from the entire sentence. 
If we further disable the Regressive Connection, the accuracy decreases as well (e.g. $2.6\%,4.6\%,6.0\%$ top-5 
further accuracy loss from the three heads). The removal of regressive connection indicates the removal of sequential dependence knowledge of pre-generated tokens, resulting in accuracy loss. 
Finally, \Sys(Baichuan) becomes Medusa(Baichuan) after further disabling Augmenting Block, losing additional $(4.3\%,8.1\%,8.9\%)$ top-5 accuracy from all heads, respectively. The missing of augmenting sequential knowledge makes it harder for succeeding to perform speculation.

Overall comparing \Sys(Baichuan) with Medusa(Baichuan), \Sys approach brings sequential knowledge from pre-generated speculative tokens as well as the input sentence, improving performance of all speculative heads, especially the latter two. The same observation is also confirmed in Figure \ref{fig:ab_ag}.

\paragraph{Exploration on Augmenting Block}
We further explore the potential variants of Augmenting Block, including: a whole transformer block (the actual architecture in \Sys(Baichuan)), attention block only , MLP block only and no Augmenting Block. 
Figure \ref{fig:ab_ag} shows accuracy of \Sys with different types of Augmenting Block equipped. The transformer block contributes $(4.9\%,9.0\%,9.4\%)$ top-5 accuracy to the heads compared with no augmenting block, in which the attention block plays the major role (increasing $1.9\%,4.4\%,5.1\%$ top-5 accuracy when only attention block enabled for Augmenting Block). While the MLP block only provides $(1.0\%,1.2\%,0.7\%)$ accuracy improvement, thus we leave it as an optional component.

The attention mechanism focuses on extracting relationships between tokens in the sentence, making it easier to learn for feature augmentation. Although the performance gain from incorporating MLP is not significant, we chose to enable it in order to improve prediction accuracy, as it has minimal time and memory overhead. 

% No Augmenting Block, Attention only, MLP only.

\section{Related Works}
\paragraph{Speculative Inference} Since Speculative decoding for LLM first proposed in \cite{pmlr-v202-leviathan23a,chen2023accelerating}, multiple optimization technologies has been studied. Tree attention was explored in \cite{miao2024specinfer} and widely applied for verifying multiple speculations in a single step. Several early works \cite{NEURIPS2023_7b97adea,liu2023online,monea2023pass,spector2023accelerating,zhou2024distillspec,zhang2023draft,hooper2024speed,chen2024cascade} studied how to improve separated draft models, and some works \cite{yang2023inference,he2024rest,fu2024break} also explored training-free draft model architecture, while more recent works such as \cite{cai2024medusa,bhendawade2024speculative,yi2024generation,du2024glide} also drew more attention to the integrated draft model. \Sys is one of the extension based on such lightweight speculator.

\paragraph{Regressive Speculator} There are some recent approaches that also explore the potential superiority of the regressive speculator. Zhang et al. \cite{zhang2024recurrent} use an MLP layer as a regressive block, and
Hydra \cite{ankner2024hydra} also introduces an additional block in their implementation. 
Eagle \cite{li2024eagle} also introduces a regressive transformer block to speculate. 
Chimera \cite{zeng2024chimera} proposed Trigram Encoder and Full Context Encoder as regressive speculators. 
The primary distinction of \Sys is the use of cross Attention Decoder and the exploration on Augmenting Block, with the aim of optimising the utilisation of sequential knowledge derived from both pre-specified tokens and the input sentence. In addition, \Sys focuses on throughput improvement at larger batch sizes and smaller tree sizes, which has not been sufficiently addressed in previous speculative decoding work.
% , as well as the discussion on token tree construction.

% \paragraph{Tree Construction} We also notice some recent works discussed about token tree construction and sampling beyond the architecture of speculator. 
% Zhang et al. \cite{} applies prefix match to the beam search results to construct token tree with regressive speculator. 
% Sequoia \cite{} theoretically analyses both the salable tree construction and token sampling algorithm, and also proposes dynamic programming tree construction algorithm similar to ours presented in Section \ref{sec:construct_dp}. The difference in our approach is selecting optimal sub-tree on given token tree with tree-based dynamic programming \CA{To Nolan: statement not strong, maybe rephrase}, and the method on conditional probabilities estimation with sequential dependency and speculators' confidence information in \Sys.

\section{Conclusion}
We present \Sys, an extension of the Medusa method that considers sequential knowledge in speculation generation.  \Sys exploits sequential knowledge from pre-generated speculative tokens (Section \ref{sec:rc}), the entire input sentence (Section \ref{sec:ab}) and their combination (Section \ref{sec:ad}), achieving $11.7\%$ - $26.4\%$ more top-5 accuracy for speculative heads and a $1.26\times$ - $1.47\times$ throughput improvement when deploying \Sys on \texttt{Baichuan Large} model compared with Medusa method (with 50\% - 76\% more speculative tokens accepted), and at most $2.56\times$ with vanilla auto-regressive decoding. The main contribution to accuracy comes from the latter heads ($+21.7\%$ - $+26.4\%$) compared to $+11.7\%$ for the first head. Such evidence support the view that the auto-regressive mechanism is an effective approach to improve the accuracy of speculation.

%%%%%%%%% BODY TEXT

%%%%%%%%% REFERENCES
\bibliographystyle{plain}
\bibliography{egbib}

\appendix
\newpage
\section{Appendix}

\subsection{More End-to-end Results on Baichuan Large model}
\label{sec:ap_e2e}
\begin{table}[htb]
    \centering
    \small
    \begin{tabular}{|P{0.9cm}|P{1.3cm}|P{1.0cm}P{1.0cm}P{1.0cm}P{1.0cm}P{1.0cm}P{1.0cm}|}
    \hline
        \multirow{2}{*}{Task} & \multirow{2}{*}{Approach}  & \multicolumn{6}{c|}{Tokens/second} \\
        &&bs=4 & bs=8 & bs=16 & bs=24 & bs=32 & bs=48\\\hline\hline
\multirow{3}{=}{\centering\textbf{RA}} & \Sys &\textbf{120.7}&\textbf{186.2}&\textbf{270.2}&\textbf{309.5}&\textbf{346.2}&\textbf{372.5}\\
        & Medusa&108.0&160.7&234.5&271.4&287.2&300.3\\
        & Vanilla&67.9&117.5&195.9&237.6&279.4&295.1\\\hline

\multirow{3}{=}{\centering\textbf{MC}} & \Sys &\textbf{121.1}&\textbf{175.9}&\textbf{262.7}&\textbf{324.4}&\textbf{383.8}&\textbf{408.9}\\
        & Medusa&101.9&152.9&222.4&282.7&303.5&324.4\\
        & Vanilla&72.1&127.4&204.4&266.9&324.6&346.2\\\hline

\multirow{3}{=}{\centering\textbf{Code}} & \Sys &\textbf{165.6}&\textbf{266.0}&\textbf{411.5}&\textbf{506.8}&\textbf{622.3}&\textbf{717.6}\\
        & Medusa&130.2&218.7&354.1&442.6&515.5&562.3\\
        & Vanilla&81.4&152.5&274.6&376.3&447.5&539.0\\\hline

\multirow{3}{=}{\centering\textbf{IP}} & \Sys &\textbf{145.6}&\textbf{240.5}&\textbf{361.9}&\textbf{467.8}&\textbf{542.8}&\textbf{650.9}\\
        & Medusa&116.4&196.9&299.9&389.4&470.0&501.3\\
        & Vanilla&74.9&145.5&268.9&348.3&416.6&531.2\\\hline

\multirow{3}{=}{\centering\textbf{CA}} & \Sys &\textbf{169.3}&\textbf{290.2}&\textbf{488.5}&\textbf{638.4}&\textbf{754.4}&\textbf{938.2}\\
        & Medusa&132.3&222.1&373.3&486.8&581.3&638.0\\
        & Vanilla&90.7&186.2&362.8&515.5&615.8&887.8\\\hline

\multirow{3}{=}{\centering\textbf{RS}} & \Sys &\textbf{194.3}&\textbf{330.0}&\textbf{526.1}&\textbf{733.3}&\textbf{837.8}&\textbf{1076.1}\\
        & Medusa&154.4&248.9&423.3&552.3&688.6&804.9\\
        & Vanilla&101.6&188.2&374.4&541.9&701.8&984.1\\\hline

\multirow{3}{=}{\centering\textbf{Math}} & \Sys &\textbf{207.3}&\textbf{342.1}&\textbf{549.5}&\textbf{715.4}&\textbf{874.3}&\textbf{1067.8}\\
        & Medusa&159.1&269.8&401.3&557.0&705.7&781.8\\
        & Vanilla&84.2&168.3&302.9&440.9&535.2&780.9\\\hline

\multirow{3}{=}{\centering\textbf{Tab}} & \Sys &\textbf{178.8}&\textbf{271.6}&\textbf{433.9}&\textbf{553.1}&\textbf{659.1}&\textbf{884.3}\\
        & Medusa&123.3&186.5&360.3&417.2&587.9&648.0\\
        & Vanilla&66.7&143.0&257.3&365.3&413.0&654.4\\\hline

\multirow{3}{=}{\centering\textbf{QA}} & \Sys &\textbf{164.9}&\textbf{271.1}&\textbf{447.2}&\textbf{573.0}&\textbf{710.0}&802.3\\
        & Medusa&127.5&213.4&349.2&457.3&565.4&623.6\\
        & Vanilla&88.6&179.9&343.2&498.2&617.2&\textbf{805.4}\\\hline

\multirow{3}{=}{\centering\textbf{Med}} & \Sys &\textbf{167.7}&\textbf{282.3}&\textbf{472.4}&\textbf{609.4}&\textbf{765.9}&\textbf{894.5}\\
        & Medusa&139.2&229.0&382.7&501.5&611.7&690.0\\
        & Vanilla&90.3&186.4&361.1&518.3&628.9&891.0\\\hline
    \end{tabular}
    \vspace{10pt}
    \caption{End-to-end throughput on \texttt{Baichuan Large} model for different tasks.}
    %  \CA{To Nolan, please give some suggestion to make this table attractive.}
    \vspace{-10pt}
\end{table}

\subsection{Token Tree Size}
\label{sec:tree_size}
\begin{figure}[htb]
    \vspace{-10pt}
    \centering
    \includegraphics[width=0.82\linewidth]{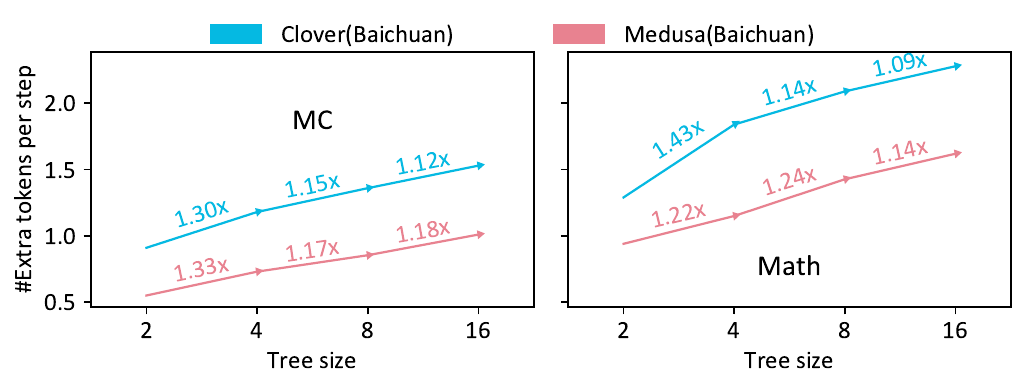}
    \caption{Extra tokens per step v.s. token tree size on \texttt{Baichuan Small} model with multi-turn conversation (\textbf{MC}) task and math(\textbf{Math}) task.}
    \label{fig:tree_size}
\end{figure}
We find marginal effects in token tree size. As the token tree size grows larger, the acceptance length is still increases but at a slower rate. Note that the horizontal axis in Figure \ref{fig:tree_size} is in exponential scale, while the vertical axis is linear. Since doubling the token tree size brings much more computation and gets closer to the computational bound, we set token tree size to 4 in our evaluation to adapt to the large batch size scenario.
%%%%%%%%%%%%%%%%%%%%%%%%%%%%%%%%%%%%%%%%%%%%%%%%%%%%%%%%%%%%

\end{document}